\documentclass{article}
\usepackage[utf8]{inputenc}
\usepackage[T1]{fontenc}
\usepackage{geometry}
\usepackage{amsfonts}
\usepackage{spconf}
\usepackage{amsmath}
\usepackage{graphicx}
\usepackage[hyphens]{url}
\usepackage{array}
\usepackage{float}
\usepackage{caption}
\usepackage[space]{grffile}
\usepackage{stmaryrd}
\usepackage{subfig}
\usepackage[section]{placeins}
\usepackage{algpseudocode}
\usepackage{color}
\usepackage{enumitem}
\usepackage{times}
\usepackage[makeroom]{cancel}
\usepackage{MnSymbol}
\usepackage{longtable}
\usepackage{multirow}

\usepackage{etoolbox}
\apptocmd{\thebibliography}{\footnotesize}{}{}
\usepackage{balance}

\usepackage{tikz}
\usetikzlibrary{bayesnet}

\tikzstyle{dot} = [circle,fill=black,draw=black,inner sep=1pt,
minimum size=5pt, node distance=0.5]

\tikzstyle{diam} = [diamond,fill=black,draw=black,inner sep=1pt,
minimum size=5pt, node distance=0.5]

\DeclareMathOperator{\diag}{diag}

\def\mbf#1{\mathbf{#1}}
\def\mbb#1{\mathbb{#1}}
\def\bs#1{\boldsymbol{#1}}

\makeatletter
\renewcommand{\paragraph}[1]{\smallskip\smallskip\noindent\textbf{#1 \hspace{.1cm}}}
\makeatother

\makeatletter
\newcounter{savesection}
\newcounter{apdxsection}
\renewcommand\appendix{\par
  \setcounter{savesection}{\value{section}}%
  \setcounter{section}{\value{apdxsection}}%
  \setcounter{subsection}{0}%
  \gdef\thesection{\@Alph\c@section}}
\newcommand\unappendix{\par
  \setcounter{apdxsection}{\value{section}}%
  \setcounter{section}{\value{savesection}}%
  \setcounter{subsection}{0}%
  \gdef\thesection{\@arabic\c@section}}
\makeatother

\setlist[itemize]{label=$\triangleright$}

\usepackage[hang,flushmargin]{footmisc}

\title{A Recurrent Variational Autoencoder for Speech Enhancement}

\name{Simon Leglaive\textsuperscript{\normalfont 1,2} \qquad Xavier Alameda-Pineda\textsuperscript{\normalfont 2}\thanks{Xavier Alameda-Pineda acknowledges the French National Research Agency (ANR) for funding the ML3RI project. This work has been partially supported by MIAI @ Grenoble Alpes, (ANR-19-P3IA-0003).} \qquad Laurent Girin\textsuperscript{\normalfont 2,3} \qquad Radu Horaud\textsuperscript{\,\normalfont 2}}

\address{\textsuperscript{1}CentraleSupélec, IETR, France \qquad \textsuperscript{2}Inria Grenoble Rh\^one-Alpes, France\\ \qquad \textsuperscript{3}Univ. Grenoble Alpes, Grenoble INP, GIPSA-lab, France}

\begin{document}
\ninept
\maketitle
\begin{abstract}
This paper presents a generative approach to speech enhancement based on a recurrent variational autoencoder (RVAE). The deep generative speech model is trained using clean speech signals only, and it is combined with a nonnegative matrix factorization noise model for speech enhancement. We propose a variational expectation-maximization algorithm where the encoder of the RVAE is fine-tuned at test time, to approximate the distribution of the latent variables given the noisy speech observations. Compared with previous approaches based on feed-forward fully-connected architectures, the proposed recurrent deep generative speech model induces a posterior temporal dynamic over the latent variables, which is shown to improve the speech enhancement results. 
\end{abstract}
\section{Introduction}
\label{sec:intro}

Speech enhancement is an important problem in audio signal processing \cite{loizou2007speech}. The objective is to recover a clean speech signal from a noisy mixture signal. 
In this work, we focus on single-channel (i.e. single-microphone) speech enhancement. 

Discriminative approaches based on deep neural networks have been extensively used for speech enhancement. They try to estimate a clean speech spectrogram or a time-frequency mask from a noisy speech spectrogram, see e.g. \cite{xu2015regression, weninger2015speech, wang2017supervised, Li_SPL_2019, Li_WASPAA2019}. Recently, deep generative speech models based on variational autoencoders (VAEs) \cite{kingma2014auto} have been investigated for single-channel \cite{bando2017statistical,Leglaive_MLSP18,Leglaive_ICASSP2019b,parienteInterspeech19} and multi-channel speech enhancement \cite{BayesianMVAE,Leglaive_ICASSP2019a,fontaine_cauchy_MVAE}. A pre-trained deep generative speech model is combined with a nonnegative matrix factorization (NMF) \cite{ISNMF} noise model whose parameters are estimated at test time, from the observation of the noisy mixture signal only. Compared with discriminative approaches, these generative methods do not require pairs of clean and noisy speech signal for training. This setting was referred to as ``semi-supervised source separation'' in previous works \cite{smaragdis2007supervised, mysore2011non, mohammadiha2013supervised}, which should not be confused with the supervised/unsupervised terminology of machine learning. 

To the best of our knowledge, the aforementioned works on VAE-based deep generative models for speech enhancement have only considered an independent modeling of the speech time frames, through the use of feed-forward and fully connected architectures. In this work, we propose a recurrent VAE (RVAE) for modeling the speech signal. The generative model is a special case of the one proposed in \cite{chung2015recurrent}, but the inference model for training is different. At test time, we develop a variational expectation-maximization algorithm (VEM) \cite{neal1998view} to perform speech enhancement. The encoder of the RVAE is fine-tuned to approximate the posterior distribution of the latent variables, given the noisy speech observations. This model induces a posterior temporal dynamic over the latent variables, which is further propagated to the speech estimate. Experimental results show that this approach outperforms its feed-forward and fully-connected counterpart.

\section{Deep generative speech model}

\subsection{Definition}
\label{subsec:VAE_speech_models}

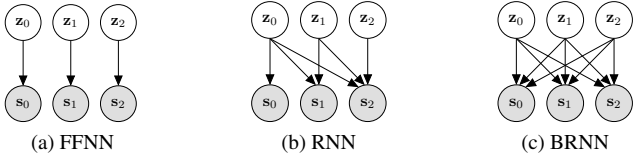
\begin{figure}[t]
\centering
\subfloat[FFNN]{
\resizebox{.22\linewidth} {!} {
\begin{tikzpicture}
\node[obs, minimum size=.8cm]                             							(s0) 	{$\mbf{s}_0$};
\node[latent, above=1cm of s0, minimum size=.8cm] 			   							(z0) 	{$\mbf{z}_0$};
\node[obs, right=.25cm of s0, minimum size=.8cm]                							(s1) 	{$\mbf{s}_1$};
\node[latent, above=1cm of s1, minimum size=.8cm] 			   							(z1) 	{$\mbf{z}_1$};
\node[obs, right=.25cm of s1, minimum size=.8cm]                							(s2) 	{$\mbf{s}_2$};
\node[latent, above=1cm of s2, minimum size=.8cm] 			   							(z2) 	{$\mbf{z}_2$};
\edge {z0} {s0} ; %
\edge {z1} {s1} ; %
\edge {z2} {s2} ; %
\end{tikzpicture}
\label{fig:FFNN_speech_model}
}
}\hfill
\subfloat[RNN]{
\resizebox{.22\linewidth} {!} {
\begin{tikzpicture}
\node[obs, minimum size=.8cm]                             							(s0) 	{$\mbf{s}_0$};
\node[latent, above=1cm of s0, minimum size=.8cm] 			   							(z0) 	{$\mbf{z}_0$};
\node[obs, right=.25 of s0, minimum size=.8cm]                							(s1) 	{$\mbf{s}_1$};
\node[latent, above=1cm of s1, minimum size=.8cm] 			   							(z1) 	{$\mbf{z}_1$};
\node[obs, right=.25 of s1, minimum size=.8cm]                							(s2) 	{$\mbf{s}_2$};
\node[latent, above=1cm of s2, minimum size=.8cm] 			   							(z2) 	{$\mbf{z}_2$};
\edge {z0.south} {s0.90} ; 
\edge {z0.south} {s1.100} ; %
\edge {z0.south} {s2.120} ; %
\edge {z1.south} {s1.90} ; %
\edge {z1.south} {s2.100} ; %
\edge {z2.south} {s2.90} ; %
\end{tikzpicture}
\label{fig:RNN_speech_model}
}
}\hfill
\subfloat[BRNN]{
\resizebox{.22\linewidth} {!} {
\begin{tikzpicture}
\node[obs, minimum size=.8cm]                             							(s0) 	{$\mbf{s}_0$};
\node[latent, above=1cm of s0, minimum size=.8cm] 			   							(z0) 	{$\mbf{z}_0$};
\node[obs, right=.25 of s0, minimum size=.8cm]                							(s1) 	{$\mbf{s}_1$};
\node[latent, above=1cm of s1, minimum size=.8cm] 			   							(z1) 	{$\mbf{z}_1$};
\node[obs, right=.25 of s1, minimum size=.8cm]                							(s2) 	{$\mbf{s}_2$};
\node[latent, above=1cm of s2, minimum size=.8cm] 			   							(z2) 	{$\mbf{z}_2$};
\edge {z0.south} {s0} ; %
\edge {z0.south} {s1.100} ; %
\edge {z0.south} {s2.120} ; %
\edge {z1.south} {s0.80} ; %
\edge {z1.south} {s1} ; %
\edge {z1.south} {s2.100} ; %
\edge {z2.south} {s0.60} ; %
\edge {z2.south} {s1.80} ; %
\edge {z2.south} {s2} ; %
\end{tikzpicture}
\label{fig:BRNN_speech_model}
}
}
\caption{Probabilistic graphical models for $N=3$.}
\label{fig:speech_models}
\end{figure}

Let $\mbf{s} = \{\mbf{s}_n \in \mathbb{C}^F \}_{n=0}^{N-1}$ denote a sequence of short-time Fourier transform (STFT) speech time frames, and $\mbf{z} = \{\mbf{z}_n \in \mathbb{R}^L \}_{n=0}^{N-1}$ a corresponding sequence of latent random vectors. We define the following hierarchical generative speech model independently for all time frames $n \in \{0,...,N-1\}$:
\begin{align}
\mbf{s}_n \mid \mbf{z} \sim \mathcal{N}_c\left(\mbf{0}, \diag\left\{ \mbf{v}_{\mbf{s},n}(\mbf{z}) \right\}\right), \hspace{.25cm} \text{with} \hspace{.25cm} \mbf{z}_n \overset{\text{i.i.d}}{\sim} \mathcal{N}\left(\mbf{0}, \mbf{I}\right),
\label{speech_generative_model}
\end{align}
and where $\mbf{v}_{\mbf{s},n}(\mbf{z}) \in \mbb{R}_+^F$ will be defined by means of a \emph{decoder} neural network. $\mathcal{N}$ denotes the multivariate Gaussian distribution for a real-valued random vector and $\mathcal{N}_c$ denotes the multivariate complex proper Gaussian distribution \cite{properComplex}. Multiple choices are possible to define the neural network corresponding to $\mbf{v}_{\mbf{s},n}(\mbf{z})$, which will lead to different probabilistic graphical models represented in Fig.~\ref{fig:speech_models}.

\paragraph{FFNN generative speech model} $\mbf{v}_{\mbf{s},n}(\mbf{z}) = \bs{\varphi}_{\text{dec}}^{\text{FFNN}}(\mbf{z}_n ; \bs{\theta}_{\text{dec}})$ where $\bs{\varphi}_{\text{dec}}^{\text{FFNN}}(\cdot \,; \bs{\theta}_{\text{dec}}) :  \mathbb{R}^L \mapsto \mathbb{R}_+^F$ denotes a feed-forward fully-connected neural network (FFNN) of parameters $\bs{\theta}_{\text{dec}}$. Such an architecture was used in \cite{bando2017statistical,Leglaive_MLSP18, Leglaive_ICASSP2019b, parienteInterspeech19, BayesianMVAE, Leglaive_ICASSP2019a,fontaine_cauchy_MVAE}. As represented in Fig.~\ref{fig:FFNN_speech_model}, this model results in the following factorization of the complete-data likelihood:
\begin{equation}
p(\mathbf{s}, \mathbf{z}; \bs{\theta}_{\text{dec}}) = \prod\nolimits_{n=0}^{N-1} p(\mbf{s}_n | \mbf{z}_n; \bs{\theta}_{\text{dec}}) p(\mbf{z}_n).
\end{equation}
Note that in this case, the speech STFT time frames are not only conditionally independent, but also marginally independent, i.e. $p(\mathbf{s}; \bs{\theta}_{\text{dec}}) = \prod\nolimits_{n=0}^{N-1} p(\mathbf{s}_n; \bs{\theta}_{\text{dec}})$.

\paragraph{RNN generative speech model}  $\mbf{v}_{\mbf{s},n}(\mbf{z}) = \bs{\varphi}_{\text{dec},n}^{\text{RNN}}(\mbf{z}_{0:n} ; \bs{\theta}_{\text{dec}})$  where $\bs{\varphi}_{\text{dec},n}^{\text{RNN}}(\cdot \,; \bs{\theta}_{\text{dec}}) :  \mathbb{R}^{L\times (n+1)} \mapsto \mathbb{R}_+^F$ denotes the output at time frame $n$ of a recurrent neural network (RNN), taking as input the sequence of latent random vectors $\mbf{z}_{0:n} = \{\mbf{z}_{n'} \in \mathbb{R}^L \}_{n'=0}^{n}$. As represented in Fig.~\ref{fig:RNN_speech_model}, we have the following factorization of the complete-data likelihood:
\begin{equation}
p(\mathbf{s}, \mathbf{z}; \bs{\theta}_{\text{dec}}) = \prod\nolimits_{n=0}^{N-1} p(\mbf{s}_n | \mbf{z}_{0:n}; \bs{\theta}_{\text{dec}}) p(\mbf{z}_n).
\end{equation}
Note that for this RNN-based model, the speech STFT time frames are still conditionally independent, \emph{but not marginally independent}.


\paragraph{BRNN generative speech model} $\mbf{v}_{\mbf{s},n}(\mbf{z}) = \bs{\varphi}_{\text{dec},n}^{\text{BRNN}}(\mbf{z} ; \bs{\theta}_{\text{dec}})$ where $\bs{\varphi}_{\text{dec},n}^{\text{BRNN}}(\cdot \,; \bs{\theta}_{\text{dec}}) :  \mathbb{R}^{L\times N} \mapsto \mathbb{R}_+^F$ denotes the output at time frame $n$ of a bidirectional RNN (BRNN) taking as input the complete sequence of latent random vectors $\mbf{z}$. As represented in Fig.~\ref{fig:BRNN_speech_model}, we end up with the following factorization of the complete-data likelihood:
\begin{equation}
p(\mathbf{s}, \mathbf{z}; \bs{\theta}_{\text{dec}}) = \prod\nolimits_{n=0}^{N-1} p(\mbf{s}_n | \mbf{z}; \bs{\theta}_{\text{dec}}) p(\mbf{z}_n).
\end{equation}
As for the RNN-based model, the speech STFT time frames are conditionally independent but not marginally.

Note that for avoiding cluttered notations, the variance $\mbf{v}_{\mbf{s},n}(\mbf{z})$ in the generative speech model \eqref{speech_generative_model} is not made explicitly dependent on the decoder network parameters $\bs{\theta}_{\text{dec}}$, but it clearly is.

\subsection{Training} 
\label{subsec:training}


We would like to estimate the decoder parameters $\bs{\theta}_{\text{dec}}$ in the maximum likelihood sense, i.e. by maximizing $\sum\nolimits_{i=1}^{I} \ln p\left(\mbf{s}^{(i)} ; \bs{\theta}_{\text{dec}}\right)$, where $\{\mbf{s}^{(i)} \in \mathbb{C}^{F \times N}\}_{i=1}^{I}$ is a training dataset consisting of $I$ i.i.d sequences of $N$ STFT speech time frames. In the following, because it simplifies the presentation, we simply omit the sum over the $I$ sequences and the associated subscript $(i)$. 

Due to the non-linear relationship between $\mbf{s}$ and $\mbf{z}$, the marginal likelihood $p(\mbf{s} ; \bs{\theta}_{\text{dec}}) = \int p(\mbf{s} | \mbf{z} ; \bs{\theta}_{\text{dec}}) p(\mbf{z}) d\mbf{z}$ is analytically intractable, and it cannot be straightforwardly optimized. We therefore resort to the framework of variational autoencoders \cite{kingma2014auto} for parameters estimation, which builds upon stochastic fixed-form variational inference \cite{Jordan1999,honkela2010approximate,Salimans2013, Hoffman2013, blei2017variational}. 

This latter methodology first introduces a variational distribution $q(\mbf{z} | \mbf{s} ; \bs{\theta}_\text{enc})$ (or inference model) parametrized by $\bs{\theta}_\text{enc}$, which is an approximation of the true intractable posterior distribution $p(\mbf{z} | \mbf{s}; \bs{\theta}_{\text{dec}})$. For any variational distribution, we have the following decomposition of log-marginal likelihood:
\begin{equation}
\ln p(\mbf{s}; \bs{\theta}_{\text{dec}}) = \mathcal{L}_{\mbf{s}}(\bs{\theta}_\text{enc}, \bs{\theta}_{\text{dec}}) + D_{\text{KL}}\big( q(\mbf{z} | \mbf{s} ; \bs{\theta}_\text{enc}) \parallel p(\mbf{z} | \mbf{s}; \bs{\theta}_{\text{dec}}) \big),
\label{log_likelihood_decomposition}
\end{equation}
where $\mathcal{L}_{\mbf{s}}(\bs{\theta}_\text{enc}, \bs{\theta}_{\text{dec}})$ is the \emph{variational free energy} (VFE) (also referred to as the evidence lower bound) defined by:
\begin{align}
\mathcal{L}_{\mbf{s}}(\bs{\theta}_\text{enc}, \bs{\theta}_{\text{dec}}) &= \mbb{E}_{q(\mbf{z} | \mbf{s} ; \bs{\theta}_\text{enc})} \left[ \ln p(\mbf{s},\mathbf{z} ; \bs{\theta}_{\text{dec}}) - \ln q(\mbf{z} | \mbf{s} ; \bs{\theta}_\text{enc}) \right] \nonumber \\
&\hspace{-1.5cm}= \mbb{E}_{q(\mbf{z} | \mbf{s} ; \bs{\theta}_\text{enc})} \left[ \ln p(\mbf{s} | \mathbf{z} ; \bs{\theta}_{\text{dec}}) \right] - D_{\text{KL}}\big( q(\mbf{z} | \mbf{s} ; \bs{\theta}_\text{enc}) \parallel p(\mbf{z} ) \big),
\label{varFreeEnergy}
\end{align}
and $D_{\text{KL}}(q \parallel p) = \mbb{E}_q[\ln q - \ln p]$ is the Kullback-Leibler (KL) divergence. As the latter is always non-negative, we see from \eqref{log_likelihood_decomposition} that the VFE is a lower bound of the intractable log-marginal likelihood. Moreover, we see that it is tight if and only if $q(\mbf{z} | \mbf{s} ; \bs{\theta}_\text{enc}) = p(\mbf{z} | \mbf{s}; \bs{\theta}_{\text{dec}})$. Therefore, our objective is now to maximize the VFE with respect to (w.r.t) both $\bs{\theta}_\text{enc}$ and $\bs{\theta}_{\text{dec}}$. But in order to fully define the VFE in \eqref{varFreeEnergy}, we have to define the form of the variational distribution $q(\mbf{z} | \mbf{s} ; \bs{\theta}_\text{enc})$.

Using the chain rule for joint distributions, the posterior distribution of the latent vectors can be exactly expressed as follows: 
\begin{equation}
p(\mathbf{z} | \mathbf{s}; \bs{\theta}_{\text{dec}}) = \prod\nolimits_{n=0}^{N-1} p(\mathbf{z}_n | \mathbf{z}_{0:n-1}, \mathbf{s} ; \bs{\theta}_{\text{dec}}),
\label{factorized_posterior_VAE}
\end{equation}
where we considered $p(\mathbf{z}_0 | \mathbf{z}_{-1}, \mathbf{s} ; \bs{\theta}_{\text{dec}}) = p(\mathbf{z}_0 | \mathbf{s} ; \bs{\theta}_{\text{dec}})$.  The variational distribution $q(\mathbf{z} | \mathbf{s}; \bs{\theta}_\text{enc})$ is naturally also expressed as:
\begin{equation}
q(\mathbf{z} | \mathbf{s}; \bs{\theta}_\text{enc}) = \prod\nolimits_{n=0}^{N-1} q(\mathbf{z}_n | \mathbf{z}_{0:n-1}, \mathbf{s} ; \bs{\theta}_\text{enc}).
\label{factorized_variational_distribution_VAE}
\end{equation}
In this work, $q(\mathbf{z}_n | \mathbf{z}_{0:n-1}, \mathbf{s} ; \bs{\theta}_\text{enc})$ denotes to the probability density function (pdf) of the following Gaussian \emph{inference model}:
\begin{equation}
\mathbf{z}_n | \mathbf{z}_{0:n-1}, \mathbf{s} \sim  \mathcal{N}\big(\bs{\mu}_{\mbf{z},n}(\mathbf{z}_{0:n-1}, \mathbf{s}), \diag\big\{ \mbf{v}_{\mbf{z},n}(\mathbf{z}_{0:n-1}, \mathbf{s}) \big\}\big),
\label{variational_distribution_VAE}
\end{equation}
where $\{ \bs{\mu}_{\mbf{z},n}, \mbf{v}_{\mbf{z},n} \}(\mathbf{z}_{0:n-1}, \mathbf{s}) \in \mbb{R}^L \times \mbb{R}_+^L$
will be defined by means of an \emph{encoder} neural network.

\paragraph{Inference model for the BRNN generative speech model} For the BRNN generative speech model, the parameters of the variational distribution in \eqref{variational_distribution_VAE} are defined by
\begin{align}
\{ \bs{\mu}_{\mbf{z},n}, \mbf{v}_{\mbf{z},n} \}(\mathbf{z}_{0:n-1}, \mathbf{s}) = \bs{\varphi}_{\text{enc},n}^{\text{BRNN}}(\mathbf{z}_{0:n-1}, \mathbf{s} ; \bs{\theta}_\text{enc}),
\end{align}
where $\bs{\varphi}_{\text{enc},n}^{\text{BRNN}}(\cdot, \cdot \,; \bs{\theta}_\text{enc}) :  \mathbb{R}^{L \times n} \times \mathbb{C}^{F \times N} \mapsto \mathbb{R}^L \times  \mathbb{R}_+^L$ denotes the output at time frame $n$ of a neural network whose parameters are denoted by $\bs{\theta}_\text{enc}$. It is composed of:
\begin{enumerate}[leftmargin=*]
\item \emph{``Prediction block''}: a causal recurrent block processing $\mbf{z}_{0:n-1}$;
\item \emph{``Observation block''}: a bidirectional recurrent block processing the complete sequence of STFT speech time frames $\mbf{s}$;
\item \emph{``Update block''}: a feed-forward fully-connected block processing the outputs at time-frame $n$ of the two previous blocks.
\end{enumerate}
If we want to sample from $q(\mathbf{z} | \mathbf{s}; \bs{\theta}_\text{enc})$ in \eqref{factorized_variational_distribution_VAE}, we have to sample recursively each $\mathbf{z}_n$, starting from $n=0$ up to $N-1$. Interestingly, the posterior is formed by running \emph{forward} over the latent vectors, and both \emph{forward} and \emph{backward} over the input sequence of STFT speech time-frames. In other words, the latent vector at a given time frame is inferred by taking into account not only the latent vectors at the previous time steps, but also all the speech STFT frames at the current, past and future time steps. The anti-causal relationships were not taken into account in the RVAE model \cite{chung2015recurrent}. 


\paragraph{Inference model for the RNN generative speech model} Using the fact that $\mathbf{z}_n$ is conditionally independent of all other nodes in Fig.~\ref{fig:RNN_speech_model} given its Markov blanket (defined as the set of parents, children and co-parents of that node) \cite{Bishop}, \eqref{factorized_posterior_VAE} can be simplified as:
\begin{equation}
p(\mathbf{z}_n | \mathbf{z}_{0:n-1}, \mathbf{s} ; \bs{\theta}_{\text{dec}}) =  p(\mathbf{z}_n | \mathbf{z}_{0:n-1}, \mathbf{s}_{n:N-1} ; \bs{\theta}_{\text{dec}}),
\end{equation}
where $\mbf{s}_{n:N-1} = \{\mbf{s}_{n'} \in \mathbb{C}^F \}_{n'=n}^{N-1}$. This conditional independence also applies to the variational distribution in \eqref{variational_distribution_VAE}, whose parameters are now given by:
\begin{align}
\{ \bs{\mu}_{\mbf{z},n}, \mbf{v}_{\mbf{z},n} \}(\mathbf{z}_{0:n-1}, \mathbf{s}) = \bs{\varphi}^{\text{RNN}}_{\text{enc},n}(\mathbf{z}_{0:n-1}, \mathbf{s}_{n:N-1} ; \bs{\theta}_\text{enc}),
\end{align}
where $\bs{\varphi}^{\text{RNN}}_{\text{enc},n}(\cdot, \cdot \,; \bs{\theta}_\text{enc}) :  \mathbb{R}^{L \times n} \times \mathbb{C}^{F \times (N-n)} \mapsto \mathbb{R}^L \times \mathbb{R}_+^L$ denotes the same neural network as for the BRNN-based model, except that the observation block is not a bidirectional recurrent block anymore, but an \emph{anti-causal} recurrent one. 
The full approximate posterior is now formed by running \emph{forward} over the latent vectors, and \emph{backward} over the input sequence of STFT speech time-frames.

\paragraph{Inference model for the FFNN generative speech model} For the same reason as before, by studying the Markov blanket of $\mathbf{z}_n$ in Fig.~\ref{fig:FFNN_speech_model}, the dependencies in \eqref{factorized_posterior_VAE} can be simplified as follows:
\begin{equation}
p(\mathbf{z}_n | \mathbf{z}_{0:n-1}, \mathbf{s} ; \bs{\theta}_{\text{dec}}) =  p(\mathbf{z}_n | \mathbf{s}_n ; \bs{\theta}_{\text{dec}}).
\end{equation}
This simplification also applies to the variational distribution in \eqref{variational_distribution_VAE}, whose parameters are now given by:
\begin{equation}
\{ \bs{\mu}_{\mbf{z},n}, \mbf{v}_{\mbf{z},n} \}(\mathbf{z}_{0:n-1}, \mathbf{s}) = \bs{\varphi}^{\text{FFNN}}_{\text{enc}}(\mbf{s}_n ; \bs{\theta}_\text{enc}),
\end{equation}
where $\bs{\varphi}^{\text{FFNN}}_{\text{enc}}(\cdot \,; \bs{\theta}_\text{enc}) :  \mathbb{C}^F \mapsto  \mathbb{R}^L \times \mathbb{R}_+^L$ denotes the output of an FFNN. Such an architecture was used in \cite{bando2017statistical,Leglaive_MLSP18, Leglaive_ICASSP2019b, parienteInterspeech19, BayesianMVAE, Leglaive_ICASSP2019a,fontaine_cauchy_MVAE}. 
This is the only case where, from the approximate posterior, we can sample all latent vectors in parallel for all time frames, without further approximation.

Here also, the mean and variance vectors in the inference model \eqref{variational_distribution_VAE} are not made explicitly dependent on the encoder network parameters $\bs{\theta}_{\text{enc}}$, but they clearly are.

\paragraph{Variational free energy} Given the generative model \eqref{speech_generative_model} and the general inference model \eqref{variational_distribution_VAE}, we can develop the VFE defined in \eqref{varFreeEnergy} as follows (derivation details are provided in Appendix~\ref{appendix:VFE}):
\begin{align}
\mathcal{L}_{\mbf{s}}(\bs{\theta}_\text{enc}, \bs{\theta}_{\text{dec}}) \overset{c}{=}& -\sum\limits_{f=0}^{F-1} \sum\limits_{n=0}^{N-1} \mbb{E}_{q(\mbf{z} | \mbf{s} ; \bs{\theta}_\text{enc})} \Big[ d_{\text{IS}} \left( |s_{fn}|^2, v_{\mbf{s},fn}(\mbf{z}) \right) \Big] \nonumber \\
&\hspace{0cm}+ \frac{1}{2}\sum\limits_{l=0}^{L-1} \sum\limits_{n=0}^{N-1} \mbb{E}_{q(\mbf{z}_{0:n-1} | \mbf{s} ; \bs{\theta}_\text{enc})} \Big[ \ln\big(v_{\mbf{z},ln}(\mathbf{z}_{0:n-1}, \mathbf{s}) \big) \nonumber \\
&\hspace{0cm}  \qquad - \mu_{\mbf{z},ln}^2(\mathbf{z}_{0:n-1}, \mathbf{s}) - v_{\mbf{z},ln}(\mathbf{z}_{0:n-1}, \mathbf{s}) \Big],
\label{varFreeEnergy_developed_IS}
\end{align}
where $\overset{c}{=}$ denotes equality up to an additive constant w.r.t $\bs{\theta}_\text{enc}$ and $\bs{\theta}_{\text{dec}}$, $d_{\text{IS}}(a, b) = a/b -\ln(a/b) - 1$ is the Itakura-Saito (IS) divergence \cite{ISNMF}, $s_{fn} \in \mbb{C}$ and $v_{\mbf{s},fn}(\mbf{z}) \in \mbb{R}_+$ denote respectively the $f$-th entries of $\mbf{s}_n$ and $\mbf{v}_{\mbf{s},n}(\mbf{z})$, and $\mu_{\mbf{z},ln}(\mathbf{z}_{0:n-1}, \mbf{s}) \in \mbb{R}$ and $v_{\mbf{z},ln}(\mathbf{z}_{0:n-1}, \mbf{s}) \in \mbb{R}_+$ denote respectively the $l$-th entry of $\bs{\mu}_{\mbf{z},n}(\mathbf{z}_{0:n-1}, \mbf{s})$ and $\mbf{v}_{\mbf{z},n}(\mathbf{z}_{0:n-1}, \mbf{s})$. 

The expectations in \eqref{varFreeEnergy_developed_IS} are analytically intractable, so we compute unbiased Monte Carlo estimates using a set  $\{ \mbf{z}^{(r)} \}_{r=1}^R$ of i.i.d. realizations drawn from $q(\mbf{z} | \mbf{s} ; \bs{\theta}_\text{enc})$. For that purpose, we use the ``reparametrization trick'' introduced in \cite{kingma2014auto}. 
The obtained objective function is differentiable w.r.t to both $\bs{\theta}_{\text{dec}}$ and $\bs{\theta}_\text{enc}$, and it can be optimized using gradient-ascent-based algorithms. 
Finally, we recall that in the final expression of the VFE, there should actually be an additional sum over the $I$ i.i.d. sequences in the training dataset $\{\mbf{s}^{(i)}\}_{i=1}^{I}$. For stochastic or mini-batch optimization algorithms, we would only consider  a subset of these training sequences for each update of the model parameters.

\section{Speech enhancement: Model and algorithm}

\subsection{Speech, noise and mixture model}

The deep generative \emph{clean} speech model along with its parameters learning procedure were defined in the previous section. For speech enhancement, we now consider a Gaussian noise model based on an NMF parametrization of the variance \cite{ISNMF}. Independently for all time frames $n \in \{0,...,N-1\}$, we have:
\begin{equation}
\mbf{b}_n \sim \mathcal{N}_c(\mbf{0}, \diag\{ \mbf{v}_{\mbf{b}, n} \}),
\end{equation}
where $\mbf{v}_{\mbf{b}, n} = (\mbf{W}_b \mbf{H}_b)_{:,n}$ with $\mbf{W}_b \in \mbb{R}_+^{F \times K}$ and $\mbf{H}_b \in \mbb{R}_+^{K \times N}$. 

The noisy mixture signal is modeled as $\mbf{x}_n = \sqrt{g_n}\mbf{s}_n + \mbf{b}_n$, where $g_n \in \mbb{R}_+$ is a gain parameter scaling the level of the speech signal at each time frame \cite{Leglaive_MLSP18}. We further consider the independence of the speech and noise signals so that the likelihood is defined by:
\begin{equation}
\mbf{x}_n \mid \mbf{z} \sim \mathcal{N}_c\left(\mbf{0}, \diag\{ \mbf{v}_{\mbf{x},n}(\mbf{z}) \} \right),
\label{mixture_likelihood}
\end{equation}
where $\mbf{v}_{\mbf{x},n}(\mbf{z}) = g_n \mbf{v}_{\mbf{s},n}(\mbf{z}) + \mbf{v}_{\mbf{b}, n}$.

\subsection{Speech enhancement algorithm}

We consider that the speech model parameters $\bs{\theta}_{\text{dec}}$ which have been learned during the training stage are fixed, so we omit them in the rest of this section. We now need to estimate the remaining model parameters $\bs{\phi} = \left\{ \mbf{g}=[g_0,...,g_{N-1}]^\top, \mbf{W}_b, \mbf{H}_b \right\}$ from the observation of the noisy mixture signal $\mbf{x} = \{\mbf{x}_n \in \mathbb{C}^F \}_{n=0}^{N-1}$. However, very similarly as for the training stage (see Section~\ref{subsec:training}), the marginal likelihood $p(\mbf{x}; \bs{\phi})$ is intractable, and we resort again to variational inference. The VFE at test time is defined by:
\begin{equation}
\mathcal{L}_{\mbf{x}}(\bs{\theta}_\text{enc}, \bs{\phi}) \hspace{-.03cm}=\hspace{-.03cm} \mbb{E}_{q(\mbf{z} | \mbf{x} ; \bs{\theta}_\text{enc})} \hspace{-.03cm}\left[\hspace{-.03cm} \ln p(\mbf{x} | \mathbf{z} ; \bs{\phi}) \hspace{-.03cm}\right] - D_{\text{KL}}\big(\hspace{-.03cm} q(\mbf{z} | \mbf{x} ; \bs{\theta}_\text{enc}) \parallel p(\mbf{z} ) \hspace{-.05cm}\big).
\label{VFE_test_time}
\end{equation}
Following a VEM algorithm \cite{neal1998view}, we will maximize this criterion alternatively w.r.t $\bs{\theta}_\text{enc}$ at the E-step, and $\bs{\phi}$ at the M-step. Note that here also, we have $\mathcal{L}_{\mbf{x}}(\bs{\theta}_\text{enc}, \bs{\phi}) \le \ln p(\mbf{x}; \bs{\phi})$ with equality if and only if $q(\mbf{z} | \mbf{x} ; \bs{\theta}_\text{enc}) = p(\mbf{z} | \mbf{x}; \bs{\phi})$.

\paragraph{Variational E-Step with fine-tuned encoder} We consider a fixed-form variational inference strategy, reusing the inference model learned during the training stage. More precisely, the variational distribution $q(\mathbf{z} | \mathbf{x}; \bs{\theta}_\text{enc})$ is defined exactly as $q(\mathbf{z} | \mathbf{s}; \bs{\theta}_\text{enc})$ in \eqref{variational_distribution_VAE} and \eqref{factorized_variational_distribution_VAE} except that $\mbf{s}$ is replaced with $\mbf{x}$. Remember that the mean and variance vectors $\bs{\mu}_{\mbf{z},n}(\cdot, \cdot)$ and $\mbf{v}_{\mbf{z},n}( \cdot, \cdot )$ in \eqref{variational_distribution_VAE} correspond to the VAE encoder network, whose parameters $\bs{\theta}_\text{enc}$ were estimated along with the parameters $\bs{\theta}_{\text{dec}}$ of the generative speech model. During the training stage, this encoder network took \emph{clean speech} signals as input. It is now \emph{fine-tuned} with a \emph{noisy speech} signal as input. For that purpose, we maximize $\mathcal{L}_{\mbf{x}}(\bs{\theta}_\text{enc}, \bs{\phi})$ w.r.t $\bs{\theta}_\text{enc}$ only, with fixed $\bs{\phi}$. This criterion takes the exact same form as \eqref{varFreeEnergy_developed_IS} except that $|s_{fn}|^2$ is replaced with $|x_{fn}|^2$ where $x_{fn} \in \mbb{C}$ denotes the $f$-th entry of $\mbf{x}_n$, $\mbf{s}$ is replaced with $\mbf{x}$, and $v_{\mbf{s},fn}(\mbf{z})$ is replaced with $v_{\mbf{x},fn}(\mbf{z})$, the $f$-th entry of $\mbf{v}_{\mbf{x},n}(\mbf{z})$ which was defined along with \eqref{mixture_likelihood}. Exactly as in Section~\ref{subsec:training}, intractable expectations are replaced with a Monte Carlo estimate and the VFE is maximized w.r.t. $\bs{\theta}_\text{enc}$ by means of gradient-based optimization techniques. In summary, we use the framework of VAEs \cite{kingma2014auto} both at training for estimating $\bs{\theta}_{\text{dec}}$ and $\bs{\theta}_\text{enc}$ from clean speech signals, and at testing for fine-tuning $\bs{\theta}_\text{enc}$ from the noisy speech signal, and with $\bs{\theta}_{\text{dec}}$ fixed. The idea of refitting the encoder was also proposed in \cite{mattei2018refit} in a different context.

\paragraph{Point-estimate E-Step} In the experiments, we will compare this variational E-step with an alternative proposed in \cite{KameokaNeuralComputation2019}, which consists in relying only on a point estimate of the latent variables. In our framework, this approach can be understood as assuming that the approximate posterior $q(\mathbf{z} | \mathbf{x}; \bs{\theta}_\text{enc})$ is a dirac delta function centered at the maximum a posteriori estimate $\mathbf{z}^\star$. Maximization of $p(\mbf{z} | \mbf{x} ; \bs{\phi}) \propto p(\mbf{x} | \mbf{z} ; \bs{\phi}) p(\mbf{z})$ w.r.t $\mbf{z}$ can be achieved by means of gradient-based techniques, where backpropagation is used to compute the gradient w.r.t. the input of the generative decoder network.

\paragraph{M-Step} For both the VEM algorithm and the point-estimate alternative, the M-Step consists in maximizing $\mathcal{L}_{\mbf{x}}(\bs{\theta}_\text{enc}, \bs{\phi})$ w.r.t. $\bs{\phi}$ under a non-negativity constraint and with $\bs{\theta}_\text{enc}$ fixed. Replacing intractable expectations with Monte Carlo estimates, the M-step can be recast as minimizing the following criterion \cite{Leglaive_MLSP18}:
\begin{align}
\mathcal{C}(\bs{\phi}) &= \sum\nolimits_{r=1}^{R} \sum\nolimits_{f=0}^{F-1} \sum\nolimits_{n=0}^{N-1} d_{\text{IS}}\left(|x_{fn}|^2, v_{\mbf{x},fn}\left(\mathbf{z}^{(r)}\right)\right),
\label{cost_M_Step}
\end{align}
where $v_{\mbf{x},fn}(\mathbf{z}^{(r)})$ implicitly depends on $\bs{\phi}$. For the VEM algorithm, $\{\mbf{z}^{(r)}\}_{r=1}^R$ is a set of i.i.d. sequences drawn from $q(\mbf{z} | \mbf{x} ; \bs{\theta}_\text{enc})$ using the current value of the parameters $\bs{\theta}_\text{enc}$. For the point estimate approach, $R=1$ and $\mbf{z}^{(1)}$ corresponds to the maximum a posteriori estimate. This optimization problem can be tackled using a majorize-minimize approach \cite{hunter2004tutorial}, which leads to the multiplicative update rules derived in \cite{Leglaive_MLSP18} using the methodology proposed in \cite{fevotte2011algorithms} (these updates are recalled in Appendix~\ref{appendix:Mstep}). 

\paragraph{Speech reconstruction} Given the estimated model parameters, we want to compute the posterior mean of the speech coefficients:
\begin{align}
\hat{s}_{fn} &= \mathbb{E}_{p(s_{fn} \mid x_{fn} ; \bs{\phi})}  [s_{fn}] = \mathbb{E}_{p(\mathbf{z} \mid \mathbf{x} ; \bs{\phi}) }\left[\frac{\sqrt{g_n}v_{\mbf{s},fn}(\mathbf{z})}{v_{\mbf{x},fn}(\mathbf{z})}\right]x_{fn}.
\label{Wiener_filtering}
\end{align}
In practice, the speech estimate is actually given by the scaled coefficients $\sqrt{g_n}\hat{s}_{fn}$. Note that \eqref{Wiener_filtering} corresponds to a Wiener-like filtering, averaged over all possible realizations of the latent variables according to their posterior distribution. As before, this expectation is intractable, but we approximate it by a Monte Carlo estimate using samples drawn from $q(\mbf{z} | \mbf{x} ; \bs{\theta}_\text{enc})$ for the VEM algorithm. For the point-estimate approach, $p(\mbf{z}|\mbf{x}; \bs{\phi})$ is approximated by a dirac delta function centered at the maximum a posteriori. 

In the case of the RNN- and BRNN-based generative speech models (see Section~\ref{subsec:VAE_speech_models}), it is important to remember that sampling from $q(\mathbf{z} | \mathbf{x}; \bs{\theta}_\text{enc})$ is actually done recursively, by sampling $q(\mathbf{z}_n | \mathbf{z}_{0:n-1}, \mathbf{x} ; \bs{\theta}_\text{enc})$ from $n=0$ to $N-1$ (see Section~\ref{subsec:training}). Therefore, there is a \emph{posterior temporal dynamic} that will be propagated from the latent vectors to the estimated speech signal, through the expectation in the Wiener-like filtering of \eqref{Wiener_filtering}. This temporal dynamic is expected to be beneficial compared with the FFNN generative speech model, where the speech estimate is built independently for all time frames.

\section{Experiments}

\paragraph{Dataset} The deep generative speech models are trained using around 25 hours of clean speech data, from the "si\_tr\_s" subset of the Wall Street Journal (WSJ0) dataset \cite{WSJ0}. Early stopping with a patience of 20 epochs is performed using the subset "si\_dt\_05" (around 2 hours of speech). We removed the trailing and leading silences for each utterance. For testing, we used around 1.5 hours of noisy speech, corresponding to 651 synthetic mixtures. The clean speech signals are taken from the "si\_et\_05" subset of WSJ0 (unseen speakers), and the noise signals from the "verification" subset of the QUT-NOISE dataset \cite{dean2015qut}. Each mixture is created by uniformly sampling a noise type among \{"café", "home", "street", "car"\} and a signal-to-noise ratio (SNR) among \{-5, 0, 5\}~dB. The intensity of each signal for creating a mixture at a given SNR is computed using the ITU-R BS.1770-4 protocol \cite{ITU}. Note that an SNR computed with this protocol is here 2.5~dB \emph{lower} (in average) than with a simple sum of the squared signal coefficients. Finally, all signals have a 16~kHz-sampling rate, and the STFT is computed using a 64-ms sine window (i.e.~$F=513$) with 75\%-overlap.

\paragraph{Network architecture and training parameters} All details regarding the encoder and decoder network architectures and their training procedure are provided in Appendix~\ref{appendix:architectures}.

\paragraph{Speech enhancement parameters} The dimension of the latent space for the deep generative speech model is fixed to $L = 16$. The rank of the NMF-based noise model is fixed to $K = 8$. $\mbf{W}_b$ and $\mbf{H}_b$ are randomly initialized (with a fixed seed to ensure fair comparisons), and $\mathbf{g}$ is initialized with an all-ones vector. For computing \eqref{cost_M_Step}, we fix the number of samples to $R=1$, which is also the case for building the Monte Carlo estimate of \eqref{Wiener_filtering}. The VEM algorithm and its "point estimate" alternative (referred to as PEEM) are run for 500 iterations. We used Adam \cite{kingma2014adam} with a step size of $10^{-2}$ for the gradient-based iterative optimization technique involved at the E-step. For the FFNN deep generative speech model, it was found that an insufficient number of gradient steps had a strong negative impact on the results, so it was fixed to 10. For the (B)RNN model, this choice had a much lesser impact so it was fixed to 1, thus limiting the computational burden.

\begin{table}[t]
	\centering
	\resizebox{.95\linewidth}{!}{
		\begin{tabular}{cc|cccc}
			Algorithm & Model & SI-SDR (dB)        & PESQ          & ESTOI         \\\hline
			MCEM \cite{Leglaive_MLSP18} & FFNN & 5.4 $\pm$ 0.4  & 2.22 $\pm$ 0.04 & 0.60 $\pm$ 0.01 \\\hline
			\multirow{3}{*}{PEEM} & FFNN & 4.4 $\pm$ 0.4  & 2.21 $\pm$ 0.04 & 0.58 $\pm$ 0.01 \\
			& RNN & 5.8 $\pm$ 0.5  & {\color{gray}\textbf{2.33}} $\pm$ 0.04 & 0.63 $\pm$ 0.01 \\
			& BRNN & 5.4 $\pm$ 0.5  & {\color{gray}\textbf{2.30}} $\pm$ 0.04 & 0.62 $\pm$ 0.01 \\\hline
			\multirow{3}{*}{VEM} & FFNN  & 4.4 $\pm$ 0.4  & 1.93 $\pm$ 0.05 & 0.53 $\pm$ 0.01 \\
			& RNN  & {\color{gray}\textbf{6.8}} $\pm$ 0.4  & {\color{gray}\textbf{2.33}} $\pm$ 0.04 & \textbf{0.67} $\pm$ 0.01 \\
			& BRNN  & \textbf{6.9} $\pm$ 0.5  & \textbf{2.35} $\pm$ 0.04 & \textbf{0.67} $\pm$ 0.01 \\\hline
			\multicolumn{2}{c|}{noisy mixture}    & -2.6 $\pm$ 0.5 & 1.82 $\pm$ 0.03 & 0.49 $\pm$ 0.01 \\
			\multicolumn{2}{c|}{oracle Wiener filtering}  & 12.1 $\pm$ 0.3 & 3.13 $\pm$ 0.02 & 0.88 $\pm$ 0.01 \\
		\end{tabular}
	}%
	\caption{Median results and confidence intervals.}
	\label{table:results}
	\vspace{-.5cm}
\end{table}

\paragraph{Results} We compare the performance of the VEM and PEEM algorithms for the three types of deep generative speech model. For the FFNN model only, we also compare with the Monte Carlo EM (MCEM) algorithm proposed in \cite{Leglaive_MLSP18} (which cannot be straightforwardly adapted to the (B)RNN model). The enhanced speech quality is evaluated in terms of scale-invariant signal-to-distortion ratio (SI-SDR) in dB \cite{LeRoux_ICASSP19}, perceptual evaluation of speech quality (PESQ) measure (between -0.5 and 4.5) \cite{rix2001perceptual} and extended short-time objective intelligibility (ESTOI) measure (between 0 and 1) \cite{taal2011algorithm}. For all measures, the higher the better. The median results for all SNRs along with their confidence interval are presented in Table~\ref{table:results}. Best results are in black-color-bold font, while gray-color-bold font indicates results that are not significantly different. As a reference, we also provide the results obtained with the noisy mixture signal as the speech estimate, and with oracle Wiener filtering. Note that oracle results are here particularly low, which shows the difficulty of the dataset. Oracle SI-SDR is for instance 7~dB lower than the one in \cite{Leglaive_MLSP18}. Therefore, the VEM and PEEM results should not be directly compared with the MCEM results provided in \cite{Leglaive_MLSP18}, but only with the ones provided here.

From Table~\ref{table:results}, we can draw the following conclusions: First, we observe that for the FFNN model, the VEM algorithm performs poorly. In this setting, the performance measures actually strongly decrease after the first 50-to-100 iterations of the algorithm. We did not observe this behavior for the (B)RNN model. We argue that the posterior temporal dynamic over the latent variables helps the VEM algorithm finding a satisfactory estimate of the overparametrized posterior model $q(\mbf{z} | \mbf{x} ; \bs{\theta}_\text{enc})$. Second, the superiority of the RNN model over the FFNN one is confirmed for all algorithms in this comparison. However, the bidirectional model (BRNN) does not perform significantly better than the unidirectional one. Third, the VEM algorithm outperforms the PEEM one, which shows the interest of using the full (approximate) posterior distribution of the latent variables and not only the maximum-a-posteriori point estimate for estimating the noise and mixture model parameters. Audio examples and code are available online \cite{companion_website}.

\vspace{-.15cm}
\section{Conclusion}
\vspace{-.15cm}

In this work, we proposed a recurrent deep generative speech model and a variational EM algorithm for speech enhancement. We showed that introducing a temporal dynamic is clearly beneficial in terms of speech enhancement. Future works include developing a Markov chain EM algorithm to measure the quality of the proposed variational approximation of the intractable true posterior distribution.

\balance
\bibliographystyle{IEEEbib_initial}
\bibliography{IEEEabrv,refs}

\begin{thebibliography}{10}

\bibitem{loizou2007speech}
P.~C. Loizou,
\newblock {\em Speech enhancement: theory and practice},
\newblock CRC press, 2007.

\bibitem{xu2015regression}
Y.~Xu, J.~Du, L.-R. Dai, and C.-H. Lee,
\newblock ``A regression approach to speech enhancement based on deep neural
  networks,''
\newblock {\em {IEEE} Trans. Audio, Speech, Language Process.}, vol. 23, no. 1,
  pp. 7--19, 2015.

\bibitem{weninger2015speech}
F.~Weninger, H.~Erdogan, S.~Watanabe, E.~Vincent, J.~Le~Roux, J.~R. Hershey,
  and B.~Schuller,
\newblock ``Speech enhancement with {LSTM} recurrent neural networks and its
  application to noise-robust {ASR},''
\newblock in {\em Proc. Int. Conf. Latent Variable Analysis and Signal
  Separation (LVA/ICA)}, 2015, pp. 91--99.

\bibitem{wang2017supervised}
D.~Wang and J.~Chen,
\newblock ``Supervised speech separation based on deep learning: An overview,''
\newblock {\em {IEEE} Trans. Audio, Speech, Language Process.}, vol. 26, no.
  10, pp. 1702--1726, 2018.

\bibitem{Li_SPL_2019}
X.~Li, S.~Leglaive, L.~Girin, and R.~Horaud,
\newblock ``Audio-noise power spectral density estimation using long short-term
  memory,''
\newblock {\em {IEEE Signal Process. Letters}}, vol. 26, no. 6, pp. 918--922,
  2019.

\bibitem{Li_WASPAA2019}
X.~Li and R.~Horaud,
\newblock ``Multichannel speech enhancement based on time-frequency masking
  using subband long short-term memory,''
\newblock in {\em Proc. IEEE Workshop Applicat. Signal Process. Audio Acoust.
  (WASPAA)}, 2019.

\bibitem{kingma2014auto}
D.~P. Kingma and M.~Welling,
\newblock ``Auto-encoding variational {Bayes},''
\newblock in {\em Proc. Int. Conf. Learning Representations (ICLR)}, 2014.

\bibitem{bando2017statistical}
Y.~Bando, M.~Mimura, K.~Itoyama, K.~Yoshii, and T.~Kawahara,
\newblock ``Statistical speech enhancement based on probabilistic integration
  of variational autoencoder and non-negative matrix factorization,''
\newblock in {\em Proc. IEEE Int. Conf. Acoust., Speech, Signal Process.
  (ICASSP)}, 2018, pp. 716--720.

\bibitem{Leglaive_MLSP18}
S.~Leglaive, L.~Girin, and R.~Horaud,
\newblock ``A variance modeling framework based on variational autoencoders for
  speech enhancement,''
\newblock in {\em Proc. IEEE Int. Workshop Machine Learning Signal Process.
  (MLSP)}, 2018, pp. 1--6.

\bibitem{Leglaive_ICASSP2019b}
S.~Leglaive, U.~\c{S}im\c{s}ekli, A.~Liutkus, L.~Girin, and R.~Horaud,
\newblock ``Speech enhancement with variational autoencoders and alpha-stable
  distributions,''
\newblock in {\em Proc. IEEE Int. Conf. Acoust., Speech, Signal Process.
  (ICASSP)}, 2019, pp. 541--545.

\bibitem{parienteInterspeech19}
M.~Pariente, A.~Deleforge, and E.~Vincent,
\newblock ``A statistically principled and computationally efficient approach
  to speech enhancement using variational autoencoders,''
\newblock in {\em Proc. Interspeech}, 2019.

\bibitem{BayesianMVAE}
K.~Sekiguchi, Y.~Bando, K.~Yoshii, and T.~Kawahara,
\newblock ``Bayesian multichannel speech enhancement with a deep speech
  prior,''
\newblock in {\em Proc. Asia-Pacific Signal and Information Processing
  Association Annual Summit and Conference (APSIPA ASC)}, 2018, pp. 1233--1239.

\bibitem{Leglaive_ICASSP2019a}
S.~Leglaive, L.~Girin, and R.~Horaud,
\newblock ``Semi-supervised multichannel speech enhancement with variational
  autoencoders and non-negative matrix factorization,''
\newblock in {\em Proc. IEEE Int. Conf. Acoust., Speech, Signal Process.
  (ICASSP)}, 2019, pp. 101--105.

\bibitem{fontaine_cauchy_MVAE}
M.~Fontaine, A.~A. Nugraha, R.~Badeau, K.~Yoshii, and A.~Liutkus,
\newblock ``{Cauchy} multichannel speech enhancement with a deep speech
  prior,''
\newblock in {\em Proc. European Signal Processing Conference (EUSIPCO)}, 2019.

\bibitem{ISNMF}
C.~F{\'e}votte, N.~Bertin, and J.-L. Durrieu,
\newblock ``{Nonnegative matrix factorization with the Itakura-Saito
  divergence: With application to music analysis},''
\newblock {\em Neural Computation}, vol. 21, no. 3, pp. 793--830, 2009.

\bibitem{smaragdis2007supervised}
P.~Smaragdis, B.~Raj, and M.~Shashanka,
\newblock ``Supervised and semi-supervised separation of sounds from
  single-channel mixtures,''
\newblock in {\em Proc. Int. Conf. Indep. Component Analysis and Signal
  Separation}, 2007, pp. 414--421.

\bibitem{mysore2011non}
G.~J. Mysore and P.~Smaragdis,
\newblock ``A non-negative approach to semi-supervised separation of speech
  from noise with the use of temporal dynamics,''
\newblock in {\em Proc. IEEE Int. Conf. Acoust., Speech, Signal Process.
  (ICASSP)}, 2011, pp. 17--20.

\bibitem{mohammadiha2013supervised}
N.~Mohammadiha, P.~Smaragdis, and A.~Leijon,
\newblock ``Supervised and unsupervised speech enhancement using nonnegative
  matrix factorization,''
\newblock {\em {IEEE} Trans. Audio, Speech, Language Process.}, vol. 21, no.
  10, pp. 2140--2151, 2013.

\bibitem{chung2015recurrent}
J.~Chung, K.~Kastner, L.~Dinh, K.~Goel, A.~C. Courville, and Y.~Bengio,
\newblock ``A recurrent latent variable model for sequential data,''
\newblock in {\em Proc. Adv. Neural Information Process. Syst. (NIPS)}, 2015,
  pp. 2980--2988.

\bibitem{neal1998view}
R.~M. Neal and G.~E. Hinton,
\newblock ``A view of the {EM} algorithm that justifies incremental, sparse,
  and other variants,''
\newblock in {\em Learning in Graphical Models}, M.~I. Jordan, Ed., pp.
  355--368. MIT Press, 1999.

\bibitem{properComplex}
F.~D. Neeser and J.~L. Massey,
\newblock ``Proper complex random processes with applications to information
  theory,''
\newblock {\em IEEE Trans. Information Theory}, vol. 39, no. 4, pp. 1293--1302,
  1993.

\bibitem{Jordan1999}
M.~I. Jordan, Z.~Ghahramani, T.~S. Jaakkola, and L.~K. Saul,
\newblock ``An introduction to variational methods for graphical models,''
\newblock {\em Machine Learning}, vol. 37, no. 2, pp. 183--233, 1999.

\bibitem{honkela2010approximate}
A.~Honkela, T.~Raiko, M.~Kuusela, M.~Tornio, and J.~Karhunen,
\newblock ``{Approximate Riemannian conjugate gradient learning for fixed-form
  variational Bayes},''
\newblock {\em Journal of Machine Learning Research}, vol. 11, no. Nov., pp.
  3235--3268, 2010.

\bibitem{Salimans2013}
T.~Salimans and D.~A. Knowles,
\newblock ``Fixed-form variational posterior approximation through stochastic
  linear regression,''
\newblock {\em Bayesian Analysis}, vol. 8, no. 4, pp. 837--882, 2013.

\bibitem{Hoffman2013}
M.~D. Hoffman, D.~M. Blei, C.~Wang, and J.~Paisley,
\newblock ``Stochastic variational inference,''
\newblock {\em Journal of Machine Learning Research}, vol. 14, no. 1, pp.
  1303--1347, 2013.

\bibitem{blei2017variational}
D.~M. Blei, A.~Kucukelbir, and J.~D. McAuliffe,
\newblock ``Variational inference: A review for statisticians,''
\newblock {\em Journal of the American Statistical Association}, vol. 112, no.
  518, pp. 859--877, 2017.

\bibitem{Bishop}
C.~M. Bishop,
\newblock {\em Pattern Recognition and Machine Learning},
\newblock Springer, 2006.

\bibitem{mattei2018refit}
P.-A. Mattei and J.~Frellsen,
\newblock ``Refit your encoder when new data comes by,''
\newblock in {\em 3rd NeurIPS workshop on Bayesian Deep Learning}, 2018.

\bibitem{KameokaNeuralComputation2019}
H.~Kameoka, L.~Li, S.~Inoue, and S.~Makino,
\newblock ``Supervised determined source separation with multichannel
  variational autoencoder,''
\newblock {\em Neural Computation}, vol. 31, no. 9, pp. 1--24, 2019.

\bibitem{hunter2004tutorial}
D.~R. Hunter and K.~Lange,
\newblock ``A tutorial on {MM} algorithms,''
\newblock {\em The American Statistician}, vol. 58, no. 1, pp. 30--37, 2004.

\bibitem{fevotte2011algorithms}
C.~F{\'e}votte and J.~Idier,
\newblock ``Algorithms for nonnegative matrix factorization with the
  $\beta$-divergence,''
\newblock {\em Neural Computation}, vol. 23, no. 9, pp. 2421--2456, 2011.

\bibitem{WSJ0}
J.~S. Garofalo, D.~Graff, D.~Paul, and D.~Pallett,
\newblock ``{CSR-I (WSJ0) Sennheiser LDC93S6B},''
  \url{https://catalog.ldc.upenn.edu/LDC93S6B}, 1993,
\newblock Philadelphia: Linguistic Data Consortium.

\bibitem{dean2015qut}
D.~B. Dean, A.~Kanagasundaram, H.~Ghaemmaghami, M.~H. Rahman, and S.~Sridharan,
\newblock ``{The QUT-NOISE-SRE protocol for the evaluation of noisy speaker
  recognition},''
\newblock in {\em Proc. Interspeech}, 2015, pp. 3456--3460.

\bibitem{ITU}
``Algorithms to measure audio programme loudness and true-peak audio level,''
\newblock {Recommendation BS.1770-4}, International Telecommunication Union
  (ITU), Oct. 2015.

\bibitem{kingma2014adam}
D.~P. Kingma and J.~Ba,
\newblock ``Adam: A method for stochastic optimization,''
\newblock in {\em Proc. Int. Conf. Learning Representations (ICLR)}, 2015.

\bibitem{LeRoux_ICASSP19}
J.~{Le Roux}, S.~{Wisdom}, H.~{Erdogan}, and J.~R. {Hershey},
\newblock ``{SDR – Half-baked or Well Done?},''
\newblock in {\em Proc. IEEE Int. Conf. Acoust., Speech, Signal Process.
  (ICASSP)}, 2019, pp. 626--630.

\bibitem{rix2001perceptual}
A.~W. Rix, J.~G. Beerends, M.~P. Hollier, and A.~P. Hekstra,
\newblock ``{Perceptual evaluation of speech quality (PESQ)-a new method for
  speech quality assessment of telephone networks and codecs},''
\newblock in {\em Proc. IEEE Int. Conf. Acoust., Speech, Signal Process.
  (ICASSP)}, 2001, pp. 749--752.

\bibitem{taal2011algorithm}
C.~H. Taal, R.~C. Hendriks, R.~Heusdens, and J.~Jensen,
\newblock ``An algorithm for intelligibility prediction of time--frequency
  weighted noisy speech,''
\newblock {\em {IEEE} Trans. Audio, Speech, Language Process.}, vol. 19, no. 7,
  pp. 2125--2136, 2011.

\bibitem{companion_website}
``Companion website,'' \url{https://sleglaive.github.io/demo-icassp2020.html}.

\bibitem{hochreiter1997long}
S.~Hochreiter and J.~Schmidhuber,
\newblock ``Long short-term memory,''
\newblock {\em Neural Computation}, vol. 9, no. 8, pp. 1735--1780, 1997.

\end{thebibliography}

\cleardoublepage
\appendix

\section{Appendix}

\subsection{Variational free energy derivation details}
\label{appendix:VFE}

In this section we give derivation details for obtaining the expression of the variational free energy in \eqref{varFreeEnergy_developed_IS}. We will develop the two terms involved in the definition of the variational free energy in \eqref{varFreeEnergy}.

\paragraph{Data-fidelity term} From the generative model defined in \eqref{speech_generative_model} we have:
\begin{align}
\mbb{E}_{q(\mbf{z} | \mbf{s} ; \bs{\theta}_\text{enc})} & \left[ \ln p(\mbf{s} | \mathbf{z} ; \bs{\theta}_{\text{dec}}) \right] \nonumber \\
=\,&  \sum_{n=0}^{N-1}  \mbb{E}_{q(\mbf{z} | \mbf{s} ; \bs{\theta}_\text{enc})} \left[ \ln p(\mbf{s}_n | \mathbf{z} ; \bs{\theta}_{\text{dec}}) \right] \nonumber \\
=\,&  -\sum_{f=0}^{F-1} \sum_{n=0}^{N-1} \mbb{E}_{q(\mbf{z} | \mbf{s} ; \bs{\theta}_\text{enc})} \left[ \ln\big(v_{\mbf{s},fn}(\mbf{z}) \big) + \frac{|s_{fn}|^2}{v_{\mbf{s},fn}(\mbf{z})} \right] \nonumber \\
& - FN\ln(\pi)
\label{derivation_data_fidelity}
\end{align}

\paragraph{Regularization term} From the inference model defined in \eqref{variational_distribution_VAE} and \eqref{factorized_variational_distribution_VAE} we have:
\begin{align}
D_{\text{KL}} & \big( q(\mbf{z} | \mbf{s} ; \bs{\theta}_\text{enc})  \parallel p(\mbf{z}) \big) \nonumber \\
=\,&  \mbb{E}_{q(\mbf{z} | \mbf{s} ; \bs{\theta}_\text{enc})} \left[ \ln q(\mbf{z} | \mbf{s} ; \bs{\theta}_\text{enc})  - \ln p(\mbf{z}) \right] \nonumber \\
=\,&  \sum_{n=0}^{N-1} \mbb{E}_{q(\mbf{z} | \mbf{s} ; \bs{\theta}_\text{enc})} \left[ \ln q(\mathbf{z}_n | \mathbf{z}_{0:n-1}, \mathbf{s} ; \bs{\theta}_\text{enc})  - \ln p(\mbf{z}_n) \right] \nonumber \\
=\,&  \sum_{n=0}^{N-1} \mbb{E}_{q(\mbf{z}_{0:n} | \mbf{s} ; \bs{\theta}_\text{enc})} \left[ \ln q(\mathbf{z}_n | \mathbf{z}_{0:n-1}, \mathbf{s} ; \bs{\theta}_\text{enc})  - \ln p(\mbf{z}_n) \right] \nonumber \\
=\,&  \sum_{n=0}^{N-1} \mbb{E}_{q(\mbf{z}_{0:n-1} | \mbf{s} ; \bs{\theta}_\text{enc})} \Big[ \mbb{E}_{q(\mbf{z}_{n} | \mbf{z}_{0:n-1}, \mbf{s} ; \bs{\theta}_\text{enc})} \big[ \ln q(\mathbf{z}_n | \mathbf{z}_{0:n-1}, \mathbf{s} ; \bs{\theta}_\text{enc}) \nonumber \\
& \hspace{1cm} - \ln p(\mbf{z}_n) \big] \Big] \nonumber \\
=\,&  \sum_{n=0}^{N-1}  \mbb{E}_{q(\mbf{z}_{0:n-1} | \mbf{s} ; \bs{\theta}_\text{enc})} \left[ D_{\text{KL}}\Big( q(\mathbf{z}_n | \mathbf{z}_{0:n-1}, \mathbf{s} ; \bs{\theta}_\text{enc})  \parallel p(\mbf{z}_n) \Big) \right]  \nonumber \\
=\,& - \frac{1}{2}\sum_{l=0}^{L-1} \sum_{n=0}^{N-1} \mbb{E}_{q(\mathbf{z}_{0:n-1} | \mbf{s} ; \bs{\theta}_\text{enc})} \Big[ \ln\big(v_{\mbf{z},ln}(\mathbf{z}_{0:n-1}, \mathbf{s}) \big) \nonumber \\
& \hspace{1cm} - \mu_{\mbf{z},ln}^2(\mathbf{z}_{0:n-1}, \mathbf{s}) - v_{\mbf{z},ln}(\mathbf{z}_{0:n-1}, \mathbf{s}) \Big] - \frac{NL}{2}.
\label{derivation_regularization}
\end{align}

Summing up \eqref{derivation_data_fidelity} and \eqref{derivation_regularization} and recognizing the IS divergence we end up with the expression of the variational free energy in \eqref{varFreeEnergy_developed_IS}.

\subsection{Update rules for the M-step}
\label{appendix:Mstep}

The multiplicative update rules for minimizing \eqref{cost_M_Step} using a majorize-minimize technique \cite{hunter2004tutorial,fevotte2011algorithms} are given by (see \cite{Leglaive_MLSP18} for derivation details):
\begin{equation}
\mathbf{H}_b \leftarrow \mathbf{H}_b \odot \left[ \frac{\mathbf{W}_b^\top \left( \mid \mathbf{X} \mid^{\odot 2} \odot \sum\limits_{r=1}^{R} \left(\mathbf{V}_\mbf{x}^{(r)} \right)^{\odot-2} \right)}{\mathbf{W}_b^\top \sum\limits_{r=1}^{R} \left(\mathbf{V}_\mbf{x}^{(r)}\right)^{\odot-1} } \right]^{\odot 1/2};
\label{updateH}
\end{equation}
\begin{equation}
\mathbf{W}_b \leftarrow \mathbf{W}_b \odot \left[ \frac{\left( \mid \mathbf{X} \mid^{\odot 2} \odot \sum\limits_{r=1}^{R} \left(\mathbf{V}_\mbf{x}^{(r)}\right)^{\odot-2} \right) \mathbf{H}_b^\top}{\sum\limits_{r=1}^{R} \left(\mathbf{V}_\mbf{x}^{(r)}\right)^{\odot-1} \mathbf{H}_b^\top } \right]^{\odot 1/2};
\label{updateW}
\end{equation}
\begin{equation}
\mathbf{g}^\top \leftarrow \mathbf{g}^\top \odot \left[ \frac{ \mathbf{1}^\top  \left[\mid \mathbf{X} \mid^{\odot 2} \odot \sum\limits_{r=1}^{R} \left(\mathbf{V}_\mbf{s}^{(r)} \odot \left(\mathbf{V}_\mbf{x}^{(r)} \right)^{\odot-2}\right)\right]}{\mathbf{1}^\top \left[ \sum\limits_{r=1}^{R} \left(\mathbf{V}_\mbf{s}^{(r)} \odot \left(\mathbf{V}_\mbf{x}^{(r)} \right)^{\odot-1}\right)\right]} \right]^{\odot 1/2},
\label{update_g}
\end{equation}
where $\odot$ denotes element-wise multiplication and exponentiation, matrix division is also element-wise, $\mathbf{V}_\mbf{s}^{(r)}, \mathbf{V}_\mbf{x}^{(r)} \in \mathbb{R}_+^{F \times N}$ are the matrices of entries $v_{\mbf{s},fn}\left(\mathbf{z}^{(r)}\right)$ and $v_{\mbf{x},fn}\left(\mathbf{z}^{(r)}\right)$ respectively, $\mathbf{X} \in \mathbb{C}^{F \times N}$ is the matrix of entries $x_{fn}$ and $\mathbf{1}$ is an all-ones column vector of dimension $F$. Note that non-negativity of $\mathbf{H}_b$, $\mathbf{W}_b$ and $\mathbf{g}$ is ensured provided that they are initialized with non-negative values.

\subsection{Neural network architectures and training}
\label{appendix:architectures}

\begin{figure}[b]
	\centering
	\subfloat[FFNN]{
		\resizebox{.20\linewidth} {!} {
			\includegraphics[width=.020\linewidth]{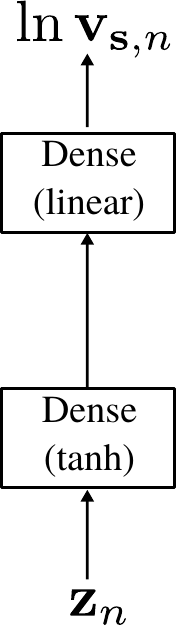}
			\label{fig:FFNN_generative_model}
		}
	}\hfill
	\subfloat[RNN]{
		\resizebox{.223\linewidth} {!} {
			\includegraphics[width=.25\linewidth]{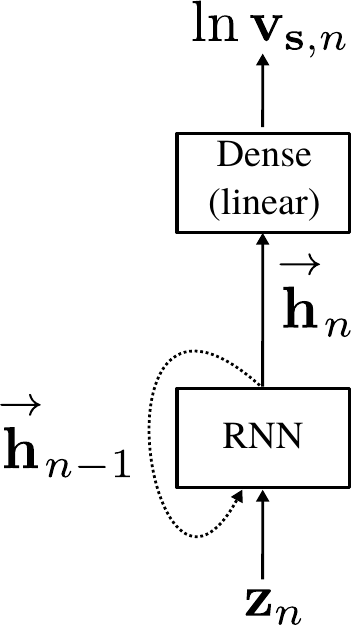}
			\label{fig:RNN_generative_model}
		}
	}\hfill
	\subfloat[BRNN]{
		\resizebox{.47\linewidth} {!} {
			\includegraphics[width=\linewidth]{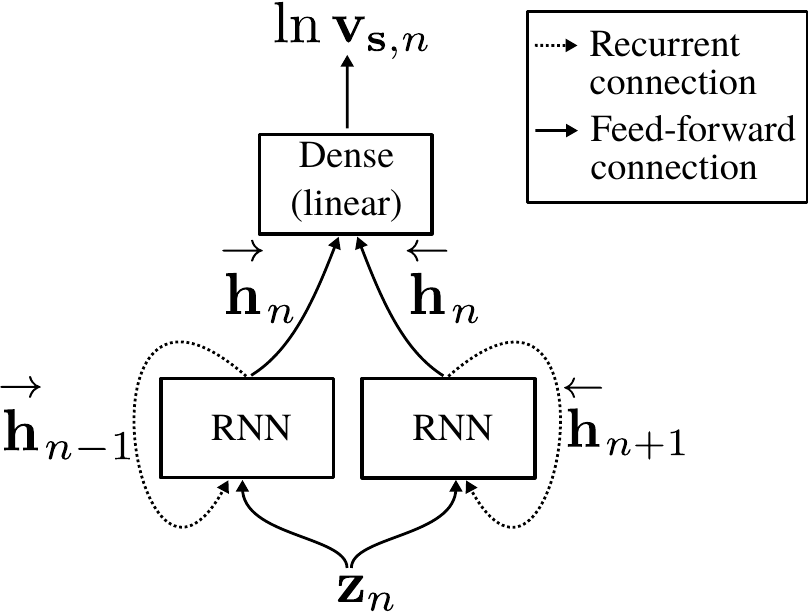}
			\label{fig:BRNN_generative_model}
		}
	}
	\caption{Decoder network architectures corresponding to the speech generative models in Fig.~\ref{fig:speech_models}.}
	\label{fig:generative_models}
\end{figure}

\begin{figure*}[!t]
	\centering
	\subfloat[BRNN]{
		\resizebox{.305\linewidth} {!} {
			\includegraphics[width=\linewidth]{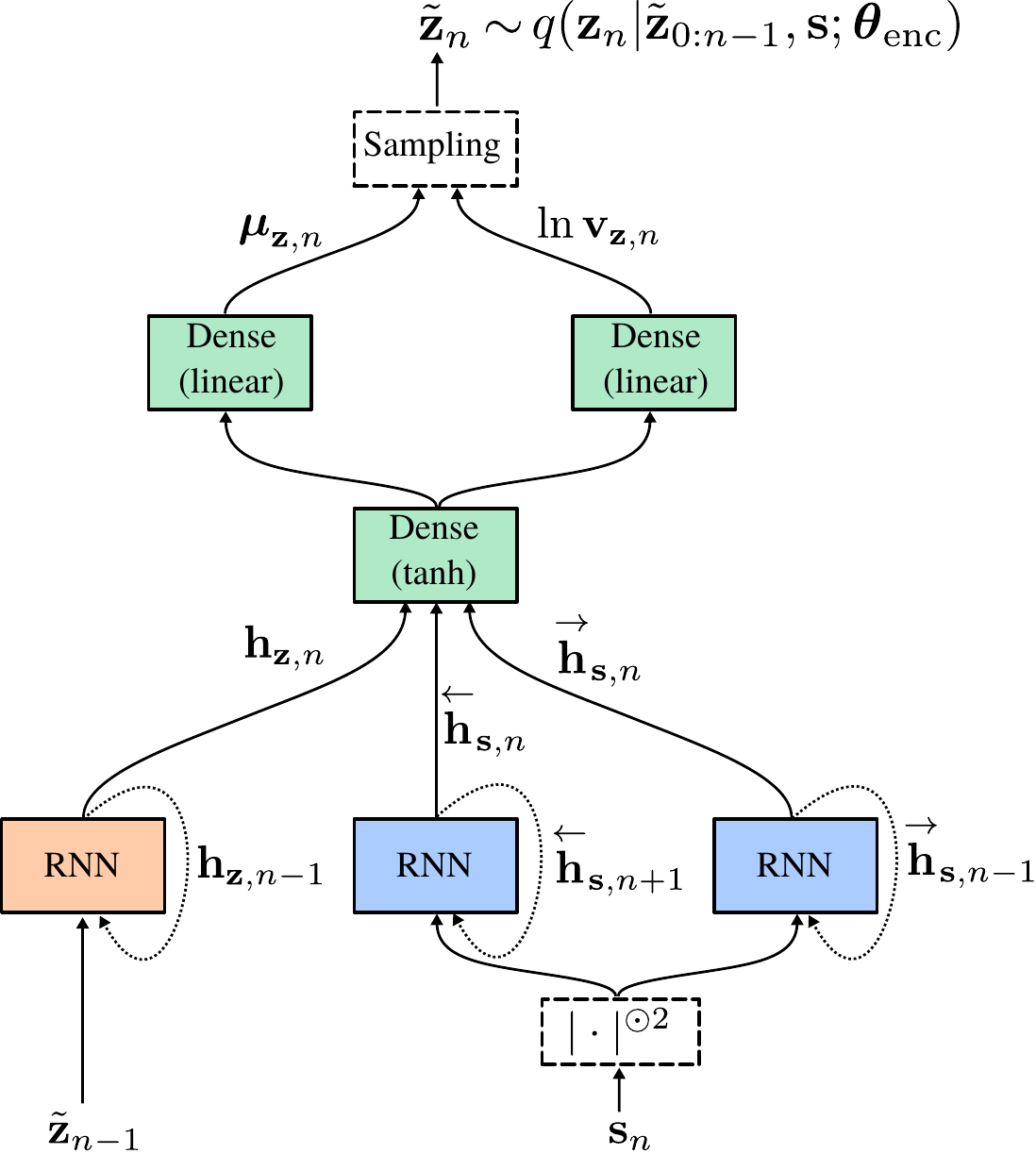}
			\label{fig:BRNN_inference_model}
		}
	}\hfill
	\subfloat[RNN]{
		\resizebox{.32\linewidth} {!} {
			\includegraphics[width=\linewidth]{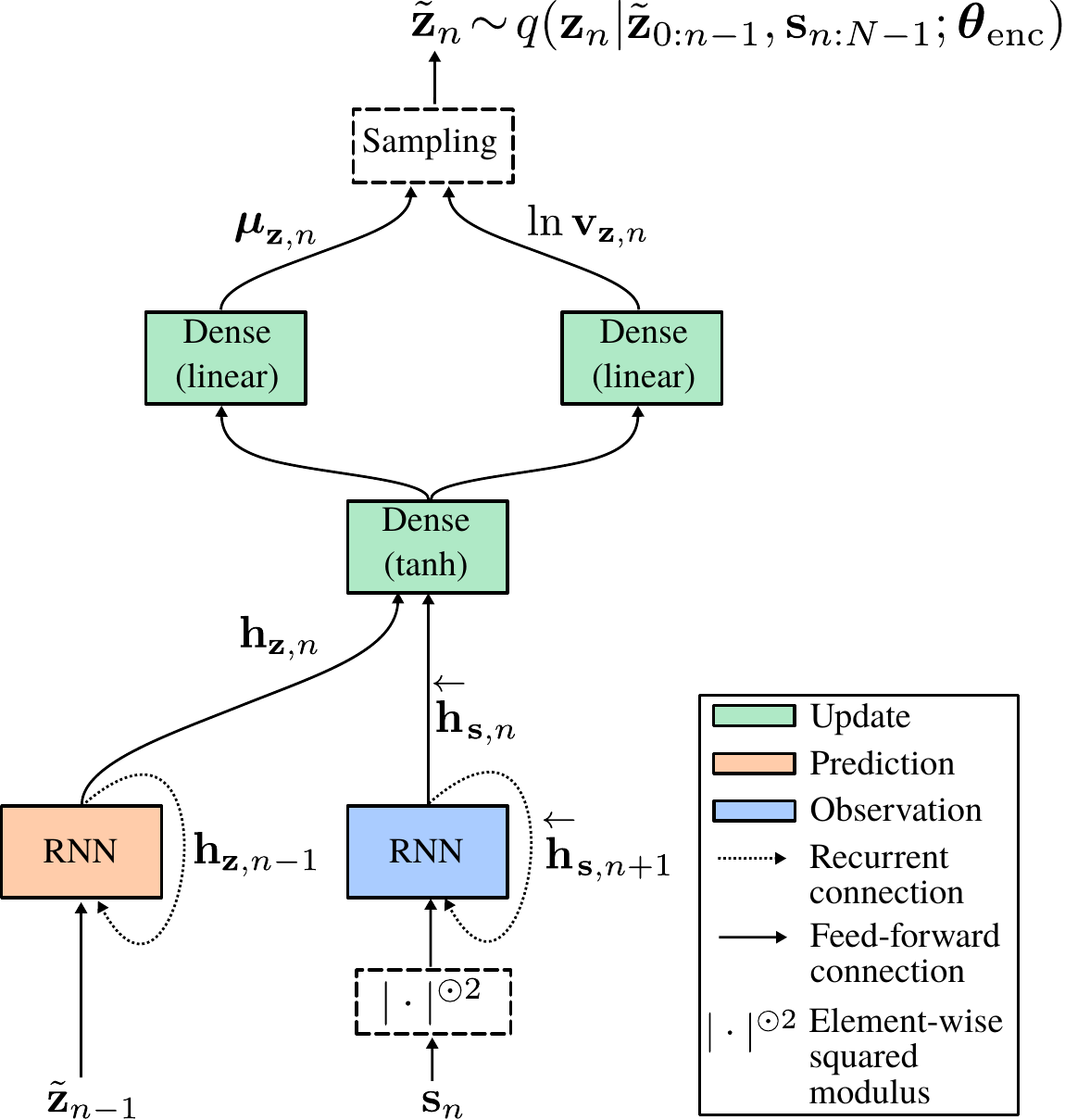}
			\label{fig:RNN_inference_model}
		}
	}\hfill
	\subfloat[FFNN]{
		\resizebox{.2\linewidth} {!} {
			\includegraphics[width=\linewidth]{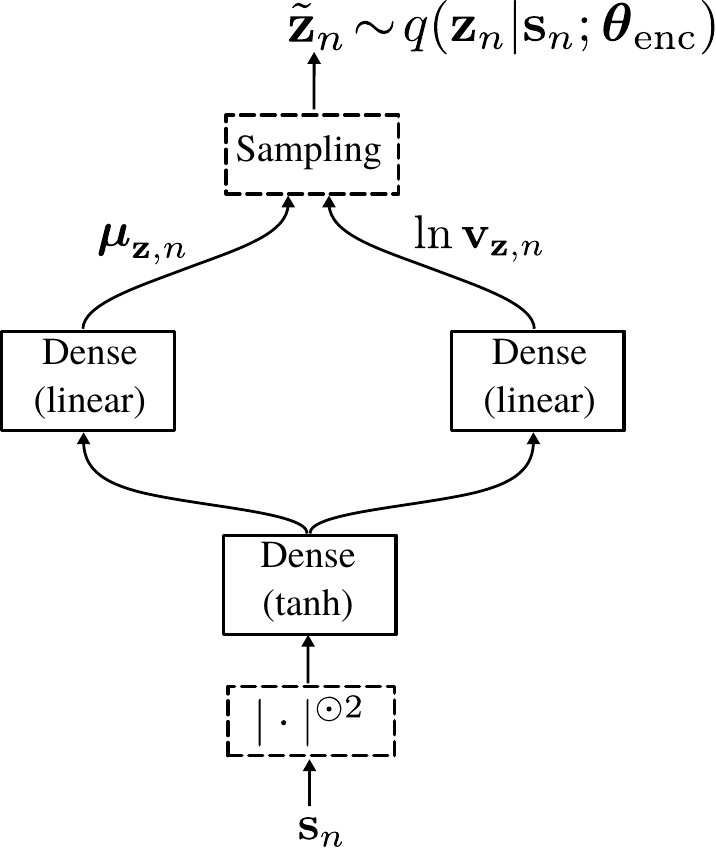}
			\label{fig:FFNN_inference_model}
		}
	}
	\caption{Encoder network architectures associated with the decoder network architectures of Fig.~\ref{fig:generative_models}.}
	\label{fig:inference_models}
\end{figure*}

The decoder and encoder network architectures are represented in Fig.~\ref{fig:generative_models} and Fig.~\ref{fig:inference_models} respectively. The "dense" (i.e. feed-forward fully-connected) output layers are of dimension $L=16$ and $F=513$ for the encoder and decoder, respectively. The dimension of all other layers was arbitrarily fixed to 128. RNN layers correspond to long short-term memory (LSTM) ones \cite{hochreiter1997long}. For the FFNN generative model, a batch is made of 128 time frames of clean speech power spectrogram. For the (B)RNN generative model, a batch is made 32 sequences of 50 time frames. Given an input sequence, all LSTM hidden states for the encoder and decoder networks are initialized to zero. For training, we use the Adam optimizer \cite{kingma2014adam} with a step size of $10^{-3}$, exponential decay rates of $0.9$ and $0.999$ for the first and second moment estimates, respectively, and an epsilon of $10^{-8}$ for preventing division by zero.

\end{document}